\pdfoutput=1

\documentclass[11pt]{article}

\usepackage{acl}

\usepackage{times}
\usepackage{latexsym}

\usepackage[T1]{fontenc}

\usepackage[utf8]{inputenc}

\usepackage{microtype}

\usepackage{balance}
\usepackage{url}
\usepackage{graphicx}
\usepackage{amsmath,amssymb,amsfonts}
\usepackage{algorithmic}
\usepackage{textcomp}
\usepackage{xcolor}
\usepackage{tikz}
\usetikzlibrary{topaths,calc}
\usepackage[ruled]{algorithm2e}
\usepackage{subfigure}
\usepackage{booktabs}
\usepackage{balance}
\usepackage{colortbl}
\usepackage{enumitem}
\usepackage{multirow}
\usepackage[normalem]{ulem}

\usepackage{etoolbox}

\DeclareMathOperator*{\argmax}{arg\,max}
\DeclareMathOperator*{\argmin}{arg\,min}

\newcommand{\revise}[1]{#1}

\title{Open Relation Modeling: Learning to Define Relations between Entities}

\author{Jie Huang$^{1}$ $\quad$ Kevin Chen-Chuan Chang$^{1}$ $\quad$ Jinjun Xiong$^{2}$ $\quad$ Wen-mei Hwu$^{1,3}$ \\
 $^1$University of Illinois at Urbana-Champaign, USA \\
 $^2$University at Buffalo, USA \\
 $^3$NVIDIA, USA \\
 \texttt{\{jeffhj, kcchang, w-hwu\}@illinois.edu} \\
 \texttt{jinjun@buffalo.edu}
}

\begin{document}
\maketitle
\begin{abstract}
Relations between entities can be represented by different instances, e.g., a sentence containing both entities or a fact in a Knowledge Graph (KG). However, these instances may not well capture the general relations between entities, may be difficult to understand by humans, even may not be found due to the incompleteness of the knowledge source. In this paper, we introduce the \textit{Open Relation Modeling} problem-- given two entities, generate a coherent sentence describing the relation between them. To solve this problem, we propose to teach machines to generate definition-like relation descriptions by letting them learn from defining entities. Specifically, we fine-tune Pre-trained Language Models (PLMs) to produce definitions conditioned on extracted entity pairs. To help PLMs reason between entities and provide additional relational knowledge to PLMs for open relation modeling, we incorporate reasoning paths in KGs and include a reasoning path selection mechanism. Experimental results show that our model can generate concise but informative relation descriptions that capture the representative characteristics of entities.\footnote{Code and data are available at \url{https://github.com/jeffhj/open-relation-modeling}.}
\end{abstract}

\section{Introduction}

\label{sec:intro}

People are always interested in relations between entities. 
To learn about a new concept, people want to know how this concept relates to the ones they are familiar with; when getting two related entities of interest, people ask how exactly they are related.

\begin{figure}[tp!]
\centerline{\includegraphics[width=\linewidth]{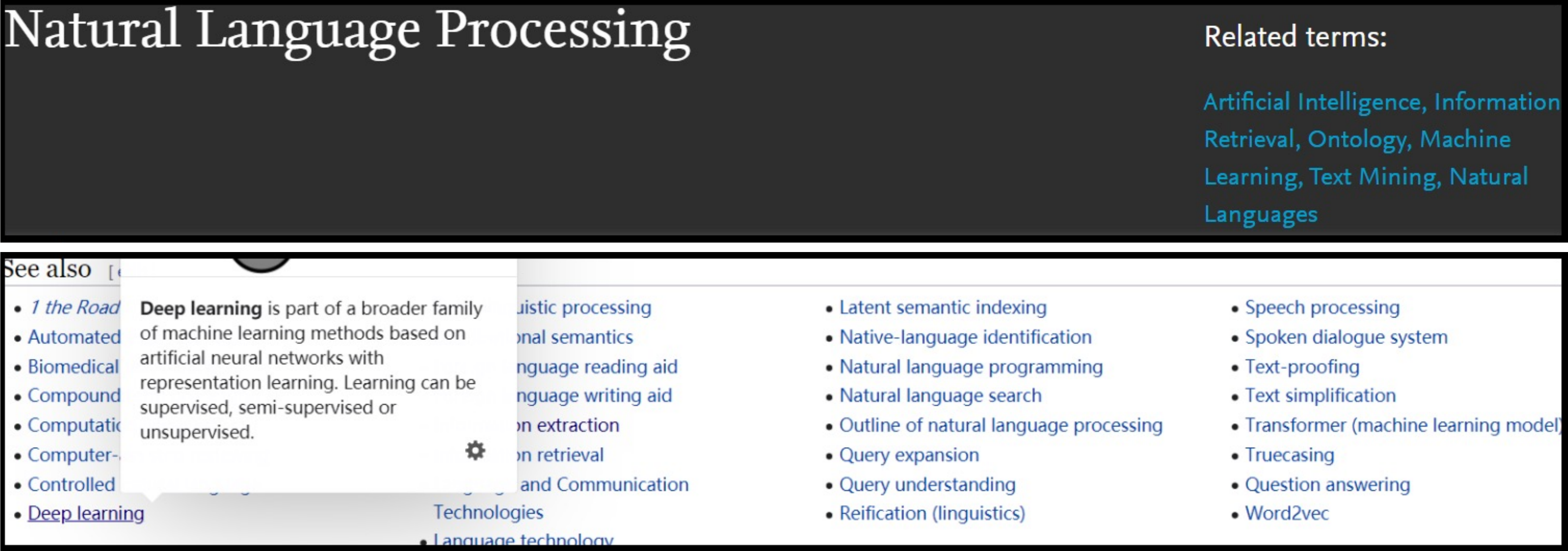}}
\caption{Examples of two current services: \textit{Elsevier's ScienceDirect Topics} (top) and \textit{Wikipedia's ``see also''} (bottom), both of which lack open relation modeling.}
\label{fig:screenshot}
\vspace{-4mm}
\end{figure}

However, although existing systems identify related entities, 
they do not provide features for exploring relations between entities. For instance, in Figure \ref{fig:screenshot}, the top is the \textit{ScienceDirect Topics} feature of Elsevier, which lists several related terms without any annotation; the bottom is the ``\textit{see also}'' feature of Wikipedia, where the annotation of \textit{deep learning} is not specific to the context of \textit{natural language processing}. 
Users cannot get how \textit{deep learning} and \textit{NLP} are related by reading the annotation,
while \textit{deep learning} is used heavily recently for \textit{NLP}.  \looseness=-1

Besides, even relations are represented, they may not be interpretable to humans.
There are different ways to represent relations between entities.
For example,
if two entities co-occur in a sentence, they are possibly related and the relation can be implied by the sentence.
From a structured perspective, 
a relation can be represented as a fact or a multi-hop reasoning path between two entities in a Knowledge Graph (KG).
However, 
for humans without too much prior knowledge about the entities, it is still difficult to understand the relations by reading them. 
For example, from sentence ``\textit{we study data mining and database.}'' or fact ``(data mining, \textit{facet of}, database)'', humans can guess \textit{data mining} and \textit{database} are related fields, but they cannot know exactly how they are related. 
Besides, due to the limited size of the corpus or the incompleteness of the KG, for many related entities, we may not extract a sentence or a fact containing both entities.

Based on the above observation, a system for exploring relations between entities needs to meet the following requirements:
1) \textbf{interpretability}: providing interpretable relation descriptions, with which humans can easily understand relations between entities;
2) \textbf{openness}: dealing with a wide range of related entities, including those neither co-occur in a corpus nor be connected in a knowledge graph, 
where types of relations are not required to be explicitly pre-specified.

To achieve a system meeting with the above requirements, we introduce a novel task-- \textbf{\emph{Open Relation Modeling}}, 
i.e., generating coherent sentences describing general relations between entities, where types of relations do not need to be pre-specified.
Different from open relation extraction, which aims to \emph{extract} relational facts between entities from an open-domain corpus \cite{banko2007open},
open relation modeling aims to \emph{generate} a concise but informative sentence, capturing the representative characteristics of the given entities and their relation.
From the perspective of interpretability,
compared to open relation extraction whose outputs are phrases with low interpretability,
e.g., \textit{(data mining methods, to be integrate within, the framework of traditional database systems)} by Ollie \cite{schmitz2012open},
open relation modeling improves the interpretability of entity relations.
For example, for \textit{data mining} and \textit{database}, we want to generate a sentence like ``\textit{\emph{data mining} is a process of extracting and discovering patterns in large data sets involving methods at the intersection of machine learning, statistics, and \emph{database} systems}.'' Such a relation description is informative and easy to understand since it contains important and precise information about entities and their relation. \looseness=-1

To solve the task, we propose to teach machines to learn from defining entities.
Definitions of entities are highly summarized sentences that capture the most representative characteristics of entities, where the general relations between the defined entity and other entities in the definitions are well captured.
Therefore, 
we suggest to \textbf{\emph{find the general relation between two entities by defining one entity in terms of the other entity}}.
To achieve this,
we first collect definitions of entities and extract entity pairs from the definitions. Then we teach machines to generate definition-like relation descriptions by training a language generation model to produce definitions of entities conditioned on extracted entity pairs.

To generate informative relation descriptions, machines need knowledge about entities and relations.
Therefore, we apply Pre-trained Language Models (PLMs) \cite{radford2019language,NEURIPS2020_1457c0d6,lewis2020bart,raffel2020exploring}, which have recently been shown to contain rich relational knowledge of entities \cite{petroni2019language,roberts2020much,wang2020language,liu2021gpt}.
To utilize knowledge to describe relations between entities, machines also need to reason between entities.
We incorporate reasoning paths in KGs to help PLMs do multi-hop reasoning and provide additional relational knowledge to PLMs. 
We also design a reasoning path selection mechanism by confidence estimation of PLMs to select interpretable and informative reasoning paths, which are then incorporated by PLMs for open relation modeling.  \looseness=-1

We conduct both quantitative and qualitative experiments.
Experimental results show that, after learning from definitions of entities, 
PLMs have a great ability to describe relations between entities concisely and informatively. 
By incorporating reasoning paths and including the reasoning path selection mechanism, machines can often generate relation descriptions well capturing relations between entities, with only minor errors that do not affect the understanding of relations.
We also conduct error analysis for the proposed methods and suggest several directions for future work.

\section{Open Relation Modeling}

\subsection{Problem Statement}

The problem of \textbf{\emph{Open Relation Modeling}} can be described as: 
given two entities $x$ and $y$, corresponding to \textit{head} and \textit{tail}, the task is to generate a coherent sentence $s$ that describes the general relation between $x$ and $y$, where types of relations do not need to be pre-specified. 
More specifically, the expected output is a concise but informative sentence that captures the representative characteristics of the entities and their relation (examples of \textit{data mining} and \textit{database} as shown in Section \ref{sec:intro}).

\subsection{Open Relation Modeling: Learning from Definitions}

\label{sec:relation_modeling}

We formulate open relation modeling as a conditional sentence generation task, i.e., generating sentences capturing general relations between entities conditioned on entity pairs. Formally, we apply the standard sequence-to-sequence formulation:
given an entity pair $(x,y)$, the probability of the output relation description $s = [w_1, \dots, w_m]$ is calculated as:   \looseness=-1
\begin{equation*}
P(s|x,y) = \prod_{i=1}^{m} P(w_i|w_0,w_1, \dots , w_{i-1},x,y),
\end{equation*}
where $w_0$ is a special start token.

To generate a sentence capturing the general relation between $x$ and $y$, 
machines need to know the semantic meanings of $x$ and $y$, 
reason between them, and learn to describe their relation in a concise but informative form.
Definitions of entities, which are highly summarized (i.e., concise but informative) sentences, capture the most representative characteristics of entities. To define an entity, other entities may be included, and the relations between the defined entity and other entities are well captured.  \looseness=-1

Therefore, we propose to \emph{teach machines to describe relations between entities by letting them learn from defining entities}.
The key idea is to \emph{find the general relation between two entities by defining one entity in terms of the other entity}.
To achieve this, we first collect definitions of entities and extract entity pairs from these definitions to form entities-definition pairs (more details are in Section \ref{sec:data}). 
After that, we teach machines to generate relation descriptions with the desired characteristics by training a language generation model to produce definitions of entities conditioned on extracted entity pairs.

With the key idea in mind, the next step is to design the generation model.
Recently, \citet{bevilacqua2020generationary} show that, by fine-tuning with context-gloss pairs, pre-trained language generation models can generate the glosses/definitions for definiendums that are not seen in the training data. 
Besides, recent studies \cite{petroni2019language,wang2020language,liu2021gpt} demonstrate that pre-trained language models contain rich relational knowledge, and such relational knowledge is essential to describing relations between entities.

Therefore, we apply pre-trained language models for open relation modeling. 
Particularly, we employ BART \cite{lewis2020bart}-- a recent transformer-based encoder-decoder model. 
In our framework, we train BART to produce the definitions of entities with extracted entity pairs as input.
Specifically, we encode the entity pair $(x,y)$ as $x; y$, 
e.g., $Haste; Germany$, and fine-tune the model to generate the corresponding sentence $s$, e.g., ``\textit{Haste is a municipality in the district of Schaumburg, in Lower Saxony, Germany}''. 
By fine-turning on the training data, the model can learn the knowledge about entities and learn to connect two entities in a coherent sentence based on its ``knowledge''. 
When given a new entity pair, the model can generate a definition-like relation description that possesses the desired characteristics.
We refer to this model as \textbf{RelationBART-Vanilla}.

\subsection{Reasoning Path-Enriched Relation Modeling}

\label{sec:rpe_rm}

While PLMs can generate coherent relation descriptions with fine-tuning on the entities-definition pairs, their ability is still limited.
Recent studies \cite{forbes2019neural,zhou2020evaluating,richardson2020does} show that it is difficult for PLMs to reason based on their knowledge.
Besides, although PLMs contain rich relational knowledge implicitly, they cannot recover all the relational knowledge in a knowledge base.

Knowledge graphs, in contrast, contain rich relational knowledge explicitly. 
Relations between entities can be represented by reasoning paths extracted from KGs directly.
A good reasoning path can guide PLMs to do multi-hop reasoning and provide additional relational knowledge to PLMs for open relation modeling.

Therefore, we want to inject relational knowledge of KGs into PLMs and incorporate reasoning paths to help PLMs reason between entities. 
We achieve this by a simple encoding scheme without changing the architecture of PLMs and re-pre-training.
Given a knowledge graph $G$, for an entity pair $(x,y)$,
if there exists a reasoning path $p(x,y) = \{x, r_1, e_1, r_2, \dots, r_{k}, y\}$ in $G$, we encode $(x, y)$ as \textit{$x$; $r_1$: $e_1$; $r_2$: $\dots$; $r_{k}$: $y$}; if not, we encode $(x, y)$ as \textit{$x$; unknown: $y$}. 
With fine-tuning on the path-sentence pairs, the model can learn to utilize the relational knowledge in a reasoning path to reason between two entities and generate a coherent sentence describing the relation between them.  \looseness=-1

However, there may exist multiple reasoning paths between two entities in a KG, while not all reasoning paths are equally helpful. 
Among the reasoning paths between two entities, the shortest one usually indicates the most direct relation. 
For example, if two entities have a direct relation in a KG, the shortest reasoning path should be a 1-hop path $p(x,y) = \{ x, r_1, y\}$. 
This path can represent a reasonable relation between two entities because this is the reason why the KG includes such a fact.
Based on this observation, formally, given an entity pair $(x,y)$, the selected reasoning path is 
\begin{equation*}
\hat{p}(x,y) = \argmin_{p(x,y) \in \mathcal{P}(x,y)} len(p(x,y)),
\end{equation*}
where $\mathcal{P}(x,y)$ is the set of reasoning paths connecting $x$ and $y$ extracted from the KG and $len(\cdot)$ is the length of the reasoning path.
We name the model trained with the shortest reasoning paths\footnote{If there exist multiple shortest paths for an entity pair, we randomly choose one.} as \textbf{RelationBART-SP}. 
To keep the presented model simple and easy to be verified, 
we leave the more complex mechanism of sampling reasoning paths as future work \cite{lao2011random,xiong2017deeppath,chen2018variational}.
In the next section, we will show that PLMs can select interpretable and informative reasoning paths automatically based on confidence estimation.  \looseness=-1

\subsection{Open Relation Modeling with Reasoning Path Selection}

\label{sec:relation_modeling_with_PS}

While shortest reasoning paths can represent the most direct relations between entities, from the perspective of human/machine understanding, these paths may not be the most interpretable and informative. 
For instance, given entity pair $(\textit{Haste}, \textit{Germany})$, with sentence description $s = $ ``\textit{Haste is a municipality in the district of Schaumburg, in Lower Saxony, Germany}'', the shortest reasoning path in Wikidata KG is $p_1 = \{\textit{Haste}, \textit{country}, \textit{Germany}\}$. This reasoning path is not \textit{\textbf{interpretable}} since we only know \textit{Haste} is in \textit{Germany}, but we have no idea whether \textit{Haste} is a municipality or a district of \textit{Germany}. However, from reasoning path $p_2 = \{$\textit{Haste$,$ located in the administrative territorial entity$,$ Schaumburg$,$ country$,$ Germany}$\}$, we can know \textit{Haste} is a smaller administrative region than \textit{Schaumburg}-- possibly a municipality.
Besides, compared to $p_1$, $p_2$ is more \textit{\textbf{informative}}. With $p_1$, to generate $s$, machines need to ``guess'' the district of \textit{Haste}. However, with $p_2$, machines can predict the district of \textit{Haste} is \textit{Schaumburg} with a high confidence.

A more interpretable and informative reasoning path can guide and help machines to generate a more reasonable and precise relation description with the desired characteristics. This is because machines can more easily reason between entities with the path and incorporate more important information from the path. 
Therefore, instead of using the shortest paths, we design a mechanism to select the most interpretable and informative reasoning paths automatically.
We achieve this by the confidence estimation of PLMs,
which is motivated by related work on machine translation and speech recognition for accessing the quality of the prediction \cite{siu1999evaluation,ueffing2007word,niehues2019modeling}.
Given an entity pair $(x,y)$, with a reasoning path $p(x,y)$, a trained model $\mathcal{M}$, and the corresponding prediction $\mathcal{M}(p(x,y))$, the confidence of the prediction can be evaluated by the posterior probability $P(\mathcal{M}(p(x,y))|p(x,y))$\footnote{We use the posterior probability of BART implemented by \texttt{fairseq}. The estimation may be further improved through calibration \cite{jiang2021can}.}.
We select the reasoning path associated with the highest confidence score:
\begin{equation*}
\hat{p}(x,y) = \argmax_{p(x,y) \in \mathcal{P}(x,y)} P(\mathcal{M}(p(x,y))|p(x,y)).
\end{equation*}

Reasoning path selection by confidence estimation is intuitive since
1) if a reasoning path is more \textit{\textbf{interpretable}}, 
which means the path is easier to convert to a precise relation description,
PLMs can ``reason'' between entities based on their knowledge with less effort; 
2) if a reasoning path is more \textit{\textbf{informative}}, which means the reasoning path provides useful relational knowledge, PLMs can incorporate such information into the prediction without guessing the necessary information. 
In both cases, the confidence of the prediction will be higher.  \looseness=-1

With the reasoning path selection mechanism,
given an entity pair $(x,y)$, the generated relation description is $\mathcal{M}(\hat{p}(x,y))$, 
where $\hat{p}(x,y)$ is the reasoning path associated with the highest confidence score.
The selected reasoning path can also serve as a support of the prediction and help users to understand the relation in a structured view.
To get the trained model $\mathcal{M}$, we can directly apply \textit{RelationBART-SP} introduced in Section \ref{sec:rpe_rm}. We name RelationBART-SP with reasoning path selection as \textbf{RelationBART-SP + PS}\footnote{We encourage the model to select relatively short paths since long paths are likely to introduce redundant information and the reasoning will not be intuitive, e.g., $\{$\textit{Haste$,$ shares border with$,$ Hohnhorst$,$ shares border with$,$ Bad Nenndorf$,$ country$,$ Germany}$\}$.} 
To make the training more robust and let PLMs learn more features from valid reasoning paths, for each entity pair, we can sample more than one reasoning path, e.g., the shortest $n$ reasoning paths with hops $\leq k$, to train the model. We refer to this model as \textbf{RelationBART-MP + PS}.

\section{Experiments}

\subsection{Dataset Construction and Analysis}

\label{sec:data}

We use Wikipedia and Wikidata \cite{vrandevcic2014wikidata} to build a benchmark dataset for open relation modeling. 

The first sentences of Wikipedia are definition-like sentences connecting different entities. For instance, the first sentence of page \textit{Deep Learning} is $s=$ ``\textit{[Deep learning] (also known as deep structured learning) is part of a broader family of [machine learning] methods based on [artificial neural networks] with [representation learning].}'' 
The head entity of this sentence is \textit{deep learning}, and there are three tail entities: \textit{machine learning}, \textit{artificial neural networks}, and \textit{representation learning}, which are linked to other pages and can be easily extracted with simple text preprocessing. Combining the head entity and the three tail entities, we can construct three entity pairs, whose expected relation descriptions are all $s$. 
The version we used is 2021-03-20 dump\footnote{\url{https://dumps.wikimedia.org/enwiki/20210320}} of English Wikipedia. For each page, we extract the plain text by WikiExtractor\footnote{\url{https://github.com/attardi/wikiextractor}} and further extract the first sentence. We randomly split entity pairs to build train/dev/test sets, where the head entities \emph{do not} overlap in each set. 

\begin{table}[tp!]
\begin{center}
\small
\begin{tabular}{c|c|c|c|c}
\toprule
 & train & dev & test & \revise{test*} \\
\hline
number & 5,434,158 & 27,431 & 55,226 & \revise{7,302} \\
\end{tabular}
\begin{tabular}{c|c|c|c|c}
\toprule
 & $1$-hop & $2$-hop & $3$-hop & $>3$-hop \\
\hline
ratio (\%) & 35.14 & 17.80 & 7.33 & 39.73 \\
\bottomrule
\end{tabular}
\caption{The statistics of the data.}
\label{table:data}
\end{center}
\vspace{-4mm}
\end{table}

To provide reasoning paths for open relation modeling, we sample part of Wikidata to build a knowledge graph. Specifically, we keep facts whose head and tail entities both appear in Wikipedia. The extracted KG contains 5,033,531 entities and 23,747,210 fact triples. The relation between two entities is considered as $k$-hop if the shortest reasoning path between them is $k$-hop.

{\flushleft \textbf{Analysis and Filtering}.}
\revise{
To assess the quality of the dataset, we randomly sample 100 examples from the test set and ask human annotators to judge whether each sentence well represents entity relationships. As a result, $87\%$ of the sentences are considered as good relation descriptions. }  \looseness=-1

\revise{
To improve the quality of evaluation,
we design a rule-based method to construct a high-quality sub-test set. Specifically, we collect dependency graph for each relation description, and calculate the \textit{dependency coverage}: the ratio of tokens covered by the shortest dependency path from the head to the tail compared to all the tokens in the sentence; and \textit{surface coverage}: the ratio of tokens between the head and the tail (including head and tail) compared to all the tokens in the sentence. 
For instance, given entity pair (\textit{Walton East}, \textit{parish}) and relation description ``\textit{Walton East is a small rural village and parish established around a church at least as early as Norman times.}'' 
The shortest dependency path from the head to the tail only contains tokens \{Walton, East, is, parish\}, so the \textit{dependency coverage} is $4/20$. And there are 9 tokens between the head and tail, so the \textit{surface coverage} is $9/20$.}

\revise{
A low \textit{dependency coverage} and \textit{surface coverage} indicate that many tokens in the sentence may not be important to characterize the relation between the head and the tail; therefore, the sentence may not be a good relation description.
We keep examples whose $(\text{dependency coverage} + \text{surface coverage})/2>0.6$. 
After filtering, $96\%$ of the sentences are judged as good relation descriptions by the human annotators.}
\revise{Here we note that while the above method filters out bad examples, it also filters out many good relationship descriptions.}
Table \ref{table:data} summarizes the statistics of the data \revise{(test* denotes the filtered sub-test set)}.

\subsection{Experimental Setup}

{\flushleft \textbf{Baselines}.}
Because our task on open relation modeling is new, there is no existing baseline for model comparison. We design the following baselines/variants for evaluation:
\begin{itemize}[noitemsep,nolistsep,leftmargin=*]
\item \textbf{DefBART}: 
Since the expected output is a definition-like sentence, the model proposed in \cite{bevilacqua2020generationary} can be applied directly, i.e., generating the definition of the head entity with the head entity as input.
We can observe the performance gain of relation modeling compared to definition modeling in terms of generating definitions and see the difference between them.
\item \textbf{RelationBART-Vanilla}: 
The vanilla version of our model introduced in Section \ref{sec:relation_modeling}. 
\item \textbf{RelationBART-SP}: 
The shortest-path version of our model introduced in Section \ref{sec:rpe_rm}.
\item \textbf{RelationBART-SP + PS}:
The shortest-path version of our model, combining with the reasoning path selection mechanism (Section \ref{sec:relation_modeling_with_PS}).
\item \textbf{RelationBART-MP + PS}:
The multiple-path version of our model, combining with the reasoning path selection mechanism (Section \ref{sec:relation_modeling_with_PS}).
\end{itemize}

Without additional notation, we apply the BART-base model and denote ``Large'' when using the BART-large model. ``w/o PT'' means the BART-base model is not pre-trained.

{\flushleft \textbf{Metrics}.}
Following existing works on text generation, we apply several widely-used metrics to automatically evaluate the performance of open relation modeling, including BLEU \cite{papineni2002bleu}, ROUGE-L \cite{lin2004rouge}, METEOR \cite{banerjee2005meteor}, and BERTScore \cite{zhang2019bertscore}. 
Among them, BLEU (BL) and ROUGE-L (R-L) are based on simple string matches, and METEOR (MT) also incorporates word stems, synonyms, and paraphrases for matching. 
These three metrics mainly focus on measuring surface similarities. 
BERTScore (BS) is based on the similarities of contextual token embeddings.
We also conduct human evaluation by asking three human annotators to assign graded values (1-4) to the sampled predictions according to Table \ref{table:rating_scale}.\footnote{Details about implementation are in Appendix \ref{app:implementation}.}

\begin{table}[tp!]
\begin{center}
\small
\setlength\tabcolsep{1.5pt} 
\begin{tabular}{l|c|c|c|c}
\toprule
& \textbf{BL} & \textbf{R-L} & \textbf{MT} & \textbf{BS} \\
\midrule 
DefBART & 20.67 & 41.82 & 18.84 & 81.56 \\
\hline
RelationBART-Vanilla (w/o PT) & 26.01 & 50.84 & 23.65 & 85.37  \\
RelationBART-SP (w/o PT) & 26.60 & 51.86 & 24.15 & 85.79 \\
RelationBART-SP (w/o PT) + PS & 27.60 & 52.70 & 24.75 & 85.99  \\
RelationBART-MP (w/o PT) + PS & \textbf{28.75} & \textbf{53.46} & \textbf{25.34} & \textbf{86.43} \\
\hline
RelationBART-Vanilla & 26.81 & 51.48 & 24.14 & 85.73 \\
RelationBART-SP & 27.78 & 52.59 & 24.79 & 86.20 \\
RelationBART-SP + PS & 28.83 & 53.48 & 25.42 & 86.40  \\
RelationBART-MP + PS & \textbf{29.51} & \textbf{53.74} & \textbf{25.64} & \textbf{86.51} \\
\hline
RelationBART-Vanilla (Large) & 27.93 & 52.10 & 24.72 & 86.03 \\
RelationBART-SP (Large) & 29.21 & 53.01 & 25.37 & 86.43 \\
RelationBART-SP (Large) + PS & \textbf{30.31} & 53.85 & \textbf{25.99} & 86.61 \\
RelationBART-MP (Large) + PS & 29.72 & \textbf{54.10} & 25.89 & \textbf{86.70}\\
\bottomrule
\end{tabular}
\caption{Results of open relation modeling \revise{on the full test set (test)}.}
\label{table:relation_modeling}
\end{center}
\vspace{-3mm}
\end{table}

\begin{table}[tp!]
\begin{center}
\small
\setlength\tabcolsep{1.5pt} 
\begin{tabular}{l|c|c|c|c}
\toprule
& \textbf{BL} & \textbf{R-L} & \textbf{MT} & \textbf{BS} \\
\midrule 
DefBART & 25.98 & 47.38 & 22.39 & 83.41 \\
\hline
RelationBART-Vanilla (w/o PT) & 34.70 & 59.57 & 28.85 & 88.01  \\
RelationBART-SP (w/o PT) & 35.48 & 60.55 & 29.40 & 88.43  \\
RelationBART-SP (w/o PT) + PS & 38.62 & 62.60 & 31.07 & 89.05  \\
RelationBART-MP (w/o PT) + PS & \textbf{40.52} & \textbf{63.73} & \textbf{32.06} & \textbf{89.53} \\
\hline
RelationBART-Vanilla & 35.45 & 59.92 & 29.33 & 88.25 \\

RelationBART-SP & 36.58 & 61.15 & 30.04 & 88.75 \\
RelationBART-SP + PS & 39.93 & 63.32 & 31.80 & 89.39  \\
RelationBART-MP + PS & \textbf{41.43} & \textbf{64.15} & \textbf{32.45} & \textbf{89.64} \\
\hline
RelationBART-Vanilla (Large) & 36.53 & 60.54 & 29.90 & 88.50 \\
RelationBART-SP (Large) & 37.65 & 61.34 & 30.57 & 88.89  \\
RelationBART-SP (Large) + PS & 41.21 & 63.56 & 32.41 & 89.53 \\
RelationBART-MP (Large) + PS & \textbf{41.46} & \textbf{64.36} & \textbf{32.62} & \textbf{89.79}\\
\bottomrule
\end{tabular}
\caption{\revise{Results of open relation modeling on the filtered test set (test*).}}
\label{table:relation_modeling_filter}
\end{center}
\vspace{-4mm}
\end{table}

\begin{table}[ht]
\begin{center}
\small
\setlength\tabcolsep{2pt} 
\begin{tabular}{l|c|c|c|c}
\toprule
\textit{hard-to-reason}  \textit{($> 3$-hop)} & \textbf{BL} & \textbf{R-L} & \textbf{MT} & \textbf{BS} \\
\midrule 
RelationBART-Vanilla & 22.99 & 47.25 & 22.21 & 84.39 \\
RelationBART-SP & \textbf{23.07} & \textbf{47.36} & \textbf{22.32} & \textbf{84.42} \\
RelationBART-SP + PS & \textbf{23.07} & \textbf{47.36} & \textbf{22.32} & \textbf{84.42} \\
RelationBART-MP + PS & 22.63 & 46.91 & 21.99 & 84.24 \\
\hline
RelationBART-Vanilla (Large) & 24.24 & \textbf{47.97} & 22.88 & \textbf{84.76} \\
RelationBART-SP (Large) & \textbf{24.50} & 47.81 & \textbf{22.90} & 84.70 \\
RelationBART-SP (Large) + PS & \textbf{24.50} & 47.81 & \textbf{22.90} & 84.70 \\
RelationBART-MP (Large) + PS & 22.92 & 47.45 & 22.34 & 84.55 \\
\midrule[0.75pt]
\textit{reasonable} \textit{($\leq 3$-hop)} & \textbf{BL} & \textbf{R-L} & \textbf{MT} & \textbf{BS} \\
\midrule 
RelationBART-Vanilla & 29.61 & 54.25 & 25.56 & 86.61 \\
RelationBART-SP & 31.24 & 56.00 & 26.62 & 87.35 \\
RelationBART-SP + PS & 33.04 & 57.48 & 27.73 & 87.70 \\
RelationBART-MP + PS & \textbf{34.52} & \textbf{58.21} & \textbf{28.36} & \textbf{87.99} \\
\hline
RelationBART-Vanilla (Large) & 30.64 & 54.81 & 26.08 & 86.86 \\
RelationBART-SP (Large) & 32.66 & 56.42 & 27.20 & 87.56 \\
RelationBART-SP (Large) + PS & 34.55 & 57.81 & 28.29 & 87.85 \\
RelationBART-MP (Large) + PS & \textbf{34.69} & \textbf{58.45} & \textbf{28.54} & \textbf{88.11} \\
\bottomrule
\end{tabular}
\caption{Results of open relation modeling for \textit{reasonable} and \textit{hard-to-reason} pairs.}
\label{table:relation_modeling_split}
\end{center}
\vspace{-4mm}
\end{table}

\subsection{Open Relation Modeling}

Tables \ref{table:relation_modeling} and \ref{table:relation_modeling_filter} summarize the experimental results of open relation modeling with the automatic metrics. We observe that RelationBART-Vanilla achieves much better performance than DefBART, which demonstrates the necessity of the tail entity in terms of generating definition-like sentences that imply relations between entities. 
Besides, RelationBART variants outperform the versions without pre-training, which indicates that knowledge stored in PLMs after pre-training is helpful for open relation modeling. 
However, the improvement is not significant, which may be because the size of our training data is large;
thus the model can learn rich knowledge about entities from definitions without pre-training. 
To verify this, we also train the model with smaller sizes of data in Appendix \ref{app:small_data}. 

Compared to RelationBART-Vanilla, the models with reasoning paths all achieve better performance, 
which demonstrates that reasoning paths can help PLMs reason between entities and provide additional relational knowledge to PLMs for open relation modeling. 
Besides, the models with reasoning path selection mechanism outperform the ones without it, which indicates PLMs can select more interpretable and informative reasoning paths based on confidence estimation, and the selected reasoning paths can guide PLMs to generate more reasonable and precise relation descriptions.

We also divide the testing examples into two groups: \textit{reasonable}, where the entities can be reasoned within $3$ hops in the Wikidata knowledge graph, and \textit{hard-to-reason}, where the entities cannot be reasoned within $3$ hops. From the results shown in Table \ref{table:relation_modeling_split}, we observe that, for the \textit{reasonable} pairs, the performance improvement is significant, while for the \textit{hard-to-reason} pairs, there is not much difference in model performance. This is because, for \textit{hard-to-reason} pairs, PLMs cannot incorporate additional relational knowledge from KGs only with encoding ``\textit{$x$; unknown: $y$}''-- which shows the training of the model is stable and the variance of the results is low.
Besides, all the models perform much better on \textit{reasonable} pairs, which indicates if two entities can be reasoned in existing KGs with fewer hops, it is easier to generate their relation descriptions with PLMs, no matter whether a reasoning path is incorporated or not.

\begin{table}[tp!]
\begin{center}
\small
\begin{tabular}{l|c}
\toprule
 & \textbf{Rating (1-4)} \\
\midrule 
RelationBART-Vanilla (Large) & 2.67 \\
RelationBART-SP (Large) & 2.82 \\
RelationBART-MP (Large) + PS & \textbf{3.01} \\
\bottomrule
\end{tabular}
\caption{Qualitative results of open relation modeling.}
\label{table:human_evaluation}
\end{center}
\vspace{-4mm}
\end{table}

{\flushleft \textbf{Qualitative Evaluation}.} 
We also perform a qualitative evaluation by asking three annotators to assign graded values to relation descriptions generated by our models according to Table \ref{table:rating_scale}. 
We randomly sample 100 \textit{reasonable} entity pairs from the test set for evaluation.
The average pairwise Cohen's kappa is 0.67, which indicates a substantial agreement (0.61-0.8) \cite{landis1977measurement}. 

From Table \ref{table:human_evaluation}, we observe the performance is satisfactory. Our best model \textit{RelationBART-MP (Large) + PS} achieves a rating of about 3, which means the model can often generate a relation description that well captures the relation, where only minor errors that do not affect the understanding of the relation are included. In addition, the qualitative evaluation results are consistent with the quantitative evaluation results in Table \ref{table:relation_modeling} and Table \ref{table:relation_modeling_split}, which validates the function of reasoning paths and reasoning path selection mechanism.

\subsection{Reasoning Path Selection}

Results in Tables \ref{table:relation_modeling}, \ref{table:relation_modeling_filter}, \ref{table:relation_modeling_split}, and \ref{table:human_evaluation} indicate machines can select better reasoning paths for open relation modeling by confidence estimation. 
We also test the quality of the selected reasoning paths from a human understanding perspective. 

We randomly select 300 entity pairs from the test set and ensure all the pairs are associated with at least two reasoning paths with hops $\leq 3$. 
For each entity pair, we randomly select 2 reasoning paths and manually label which one is more interpretable and informative, i.e., humans can understand the relation between two entities more easily by reading the reasoning path. We skip pairs that are difficult to judge which path is better. 
Among the 300 pairs, 106 pairs were skipped.

Table \ref{table:reasoning_path_selection} reports the results of reasoning path selection with different methods. 
The \textit{Random Walk} baseline selects the reasoning path by the probability of generating the path starting from the head entity, which is suggested by \cite{lao2011random}. 
The \textit{Shortest Path} baseline selects the path with a shorter length (for 52 cases where the length of two paths is the same, we randomly choose one).

We can see the performance of RelationBART-MP (Large) is quite impressive, where machines make the same choices as humans in more than 80\% of the cases.
In addition, results in Table \ref{table:reasoning_path_selection} are consistent with results in Table \ref{table:relation_modeling}, 
which indicates a better reasoning path selection mechanism can promote machines to generate better relation descriptions. 

\begin{table}[tp!]
\begin{center}
\small
\begin{tabular}{l|c}
\toprule
 & \textbf{Accuracy (\%)} \\
\midrule 
Random Walk & 64.43 \\
Shortest Path & 61.34 \\
RelationBART-SP (Large) & 72.68 \\
RelationBART-MP (Large) & \textbf{80.93} \\
\bottomrule
\end{tabular}
\caption{Results of reasoning path selection.}
\vspace{-4mm}
\label{table:reasoning_path_selection}
\end{center}
\end{table}

\subsection{Generation Examples and Error Analysis}

\begin{table*}[ht]
\tiny
\setlength\tabcolsep{1.5pt} 
\begin{tabular}{p{0.15\linewidth}|p{0.29\linewidth}|p{0.49\linewidth}|c}
\toprule
\textbf{Input} & \textbf{Reasoning Path (Encoding)} & \textbf{Output} & \textbf{Score} \\
\midrule[0.75pt]
Romeries \textbf{[reference]} & \centering{-} & Romeries is a commune in the Nord department in northern France. & - \\
\hline
\multirow{6}{*}{(Romeries, France)} & Romeries; country: France \textbf{[NOT selected]} & Romeries is a commune in the Somme department in Hauts-de-France in northern France. & -0.29 \\
\cline{2-4}
& Romeries; located in the administrative territorial entity: Nord; country: France \textbf{[selected]} & Romeries is a commune in the Nord department in northern France. & -0.13 \\
\cline{2-4}
& Romeries; shares border with: Solesmes; country: France \textbf{[NOT selected]} & Romeries is a commune in the Marne department in north-eastern France. & -0.31 \\
\midrule[0.75pt]
data mining \textbf{[reference]} & \centering{-} & Data mining is a process of extracting and discovering patterns in large data sets involving methods at the intersection of machine learning, statistics, and database systems. & - \\
\hline
(data mining, artificial intelligence) & data mining; facet of: machine learning; subclass of: artificial intelligence & Data mining is a subfield of machine learning and artificial intelligence concerned with the collection, processing, and analysis of large amounts of data. & -0.95 \\
\hline
(data mining, database) & data mining; facet of: database & Data mining is the process of extracting information from a data set, such as a database. & -0.94 \\
\hline
(data mining, statistics) & data mining; facet of: machine learning; partially coincident with: statistics & Data mining is the process of collecting, processing, and analyzing large amounts of data in the form of statistics and machine learning. & -1.04 \\
\hline
(data mining, data) & data mining; facet of: machine learning; partially coincident with: statistics; studies: data & Data mining is the process of extracting useful information from data. & -0.81 \\
\midrule[0.75pt]
evaluation \textbf{[reference]} & \centering{-} & Evaluation is a systematic determination of a subject's merit, worth and significance, using criteria governed by a set of standards. & - \\
\hline
(evaluation, machine learning) & evaluation; unknown: machine learning & In computer science, evaluation is the process of evaluating a machine learning algorithm to determine whether the algorithm is performing well. & -1.30 \\
\hline
(evaluation, algorithm) & evaluation; unknown: algorithm & In computer science, evaluation is the process of evaluating an algorithm to determine whether it is correct. & -1.13 \\
\hline
(evaluation, robotics) & evaluation; unknown: robotics & In robotics, evaluation is the process of determining whether or not a particular component of a system is working properly. & -1.54 \\
\hline
(evaluation, software engineering) & evaluation; unknown: software engineering & In computer science and software engineering, evaluation is the process of determining whether a particular feature or feature should be added to a product or service. & -1.26 \\
\hline
(evaluation, computer security) & evaluation; unknown: computer security & In computer security, evaluation is the process of determining the security of a computer system. & -1.09 \\
\midrule[0.75pt]
The Association for Computational Linguistics \textbf{[reference]} & \centering{-} & The Association for Computational Linguistics (ACL) is the international scientific and professional society for people working on problems involving natural language and computation. & - \\
\hline
(The Association for Computational Linguistics, natural language processing) & The Association for Computational Linguistics; unknown: natural language processing & The Association for Computational Linguistics (ACL) is a professional association in the field of natural language processing (NLP). & -0.60 \\
\hline
(The Association for Computational Linguistics, artificial intelligence) & The Association for Computational Linguistics; unknown: artificial intelligence & The Association for Computational Linguistics (ACL) is a professional association for linguists working in the field of computational linguistics, including artificial intelligence, machine learning, natural language processing, and computational linguistics. & -0.67 \\
\bottomrule
\end{tabular}
\vspace{-1mm}
\caption{Sample of relation descriptions generated by RelationBART-MP (Large).}
\label{table:example}
\vspace{-4mm}
\end{table*}

Table \ref{table:example} shows some generation examples via the RelationBART-MP (Large) model. 
The first row of each group is the reference definition in Wikipedia.
1) The first group contains reasoning paths connecting \textit{Romeries} and \textit{France}, we can see the second reasoning path is the most interpretable and informative, where the confidence score is the highest, and the generated relation description is the same as the ground truth. 
2) The second group contains four entity pairs whose head entity is \textit{data mining}, and the reasoning path in each row is the one associated with the highest confidence score. 
By reading these generated relation descriptions, humans can better understand \textit{data mining} and its relationship with other terms.
3) The third group contains five \textit{hard-to-reason} entity pairs whose head entity is \textit{evaluation}. 
We can see the reference definition of \textit{evaluation} is quite abstract that cannot capture the relation between \textit{evaluation} and a specific field,
while by reading the generated ones, humans can understand what \textit{evaluation} means in different fields and how it relates to them.

{\flushleft \textbf{Error Analysis}.}
To further understand the quality of the outputs produced by our model and identify the remaining challenges, we investigate the error cases found by examining the generated relation descriptions. 
As a result, we found most errors can refer to as \emph{hallucinations}, i.e., producing irrelevant or contradicted facts. 
This type of error is mainly due to knowledge coming from pre-training, fine-tuning, and reasoning paths is not sufficient. 

Taking entity pair $(\textit{Romeries}, \textit{France})$ in Table \ref{table:example} as an example, if the model takes the shortest reasoning path, i.e., \textit{Romeries; country: France}, as input, a relation description that wrongly predicts the department of \textit{Romeries} will be generated. This is because knowledge about the department is missing from the reasoning path, and such detailed knowledge is also difficult to obtain from the parameters of the trained model. 

Another example is $(\textit{Play It Loud}, \textit{rock music})$, where the reference relation description is ``\textit{Play It Loud is the second studio album by the British rock group Slade.}'' The reasoning path selected by RelationBART-MP (Large) is 
$\{$\textit{Play It Loud$,$ performer$,$ Slade$,$ genre$,$ hard rock$,$ subclass of$,$ rock music}$\}$.
This reasoning path contains detailed knowledge about the performer; however, it is still difficult to judge whether \textit{Play It Loud} is a song or an album. As a result, the model generates ``\textit{Play It Loud is a song by the British rock band Slade.}'' 

Hallucination is a common issue and challenging problem in text generation. 
From the results in Table \ref{table:human_evaluation} and the generation examples, we can observe hallucination is reduced by incorporating reasoning paths and the reasoning path selection mechanism.
How to further alleviate it for open relation modeling will be our further work direction.
We discuss some possible solutions in Section \ref{sec:discussion}.

\section{Discussion}

\label{sec:discussion}

{\flushleft \textbf{Limitation of Definitional Sentences}.} 
Although a considerable number of relations can be well captured by definitional sentences, there are types of relations that are not natural to be represented by definitional sentences. 
For instance, for \emph{Kobe Bryant} and \emph{Shaq O'neal} (both are NBA players in Los Angeles Lakers), it is not natural to assume one would appear in the other's definition.
In this case, we can include a third related entity to help users to understand their relation.
For example, we can include \emph{Los Angeles Lakers} (which can be found from a knowledge graph or a corpus);
and then, we can generate two sentences: 1) ``Kobe Bryant was an NBA player in Los Angeles Lakers''; 2) ``Shaq O'neal was an NBA player in Los Angeles Lakers''. With these two sentences, users can easily understand their relation. It is also possible to design a model to synthesize these two sentences to one \cite{becker2021reconstructing}, e.g., ``Kobe Bryant and Shaq O'neal were both NBA players played in Los Angeles Lakers''. 
We leave a comprehensive solution to solve this limitation as future work.

{\flushleft \textbf{Open Relation Modeling with Diversity}.}
In the real world, multiple important relations can be associated with one entity pair. Considering this, as future work, we may generate diverse relation descriptions for one entity pair with different reasoning paths selected.

{\flushleft \textbf{Open Relation Modeling with More Knowledge}.}
Open relation modeling is a knowledge-intensive task \cite{NEURIPS2020_6b493230}, where knowledge about entities and relations is essential to solving this task. In this work, we incorporate knowledge from model pre-training, definitions of entities, and reasoning paths. 
The proposed model can achieve impressive performance, especially for reasonable entity pairs. As future work, we can leverage more external information of entities, e.g., sentences/paragraphs containing the target entities from corpora, to provide more knowledge for open relation modeling. \looseness=-1

\section{Related Work}

Previously, \citet{voskarides2015learning} study the problem of extracting sentences that describe relations between entities with direct relations in a knowledge graph. They model this task as a learning to rank problem and design a supervised learning model with manually annotated sentences. As follow-up work, \citet{huang2017learning} solve this task with training data built by leveraging clickthrough data from Web search, and \citet{voskarides2017generating} generate the description of a relationship instance in a knowledge graph by filling created sentence templates with appropriate entities. The ability of these models is limited since they heavily rely on features of entities and relations; thus these models can only handle entities with several pre-specified types (only 10 in \cite{voskarides2017generating}) of explicit relations in KGs (e.g., \textit{isMemberOfMusicGroup}), while our methods can deal with a large number of types of relations, including implicit ones (e.g., \textit{evaluation} and \textit{algorithm}), i.e., in an ``open'' setting. \looseness=-1

Recently, \citet{lin2020commongen,liu2021kg} study a constrained text generation problem that aims to generate coherent sentences describing everyday scenarios containing the given common concepts. 
Different from them, we aim to generate sentences that can explain the relation between entities intuitively and explicitly.
\citet{dognin2020dualtkb,agarwal2021knowledge} study the data-to-text generation problem \cite{kukich1983design} that converts the KG into natural text with language models. 
The focus of these works is to convert knowledge graphs into natural language, while we propose to discover relation descriptions between entities with pre-trained language models.
Besides, only common concepts or entities with direct relations are studied in these works, 
while our methods deal with entities with multi-hop relations, even including entities that cannot be reasoned in existing KGs.

\section{Conclusion}

In this paper, we introduce and study the novel \emph{open relation modeling} problem-- generating coherent sentences describing general relations between entities, where the relations can be multi-hop, even cannot be reasoned in an existing KG. 
We achieve this by teaching PLMs to learn from defining entities and select/utilize reasoning paths.
We believe this work will open a door for modeling relations between entities.
As for future work, we plan to improve our model as discussed in Section \ref{sec:discussion} and apply our methods to downstream applications, e.g., a system for users to explore relations between entities, \revise{which can be further applied to explore a taxonomy or ontology.}
We can also use the generated relation descriptions to help some related tasks, such as relation extraction \cite{bach2007review}, knowledge graph construction and completion \cite{ji2021survey}. 
The trained models can be further fine-tuned for open relation modeling on specific domains. \looseness=-1

\section*{Acknowledgements}

We thank the reviewers for their constructive feedback.
This material is based upon work supported by the National Science Foundation IIS 16-19302 and IIS 16-33755, Zhejiang University ZJU Research 083650, IBM-Illinois Center for Cognitive Computing Systems Research (C3SR)-- a research collaboration as part of the IBM Cognitive Horizon Network, grants from eBay and Microsoft Azure, UIUC OVCR CCIL Planning Grant 434S34, UIUC CSBS Small Grant 434C8U, and UIUC New Frontiers Initiative. Any opinions, findings, and conclusions or recommendations expressed in this publication are those of the author(s) and do not necessarily reflect the views of the funding agencies. \looseness=-1

\bibliography{anthology,custom}
\bibliographystyle{acl_natbib}

\clearpage

\appendix

\begin{table}[tp!]
\small
\setlength\tabcolsep{1.5pt} 
\begin{tabular}{c|p{0.85\linewidth}}
\toprule
\textbf{Rating} & \multicolumn{1}{c}{\textbf{Criterion}} \\
\midrule[0.75pt]
4 & The relation is well captured, and important information about entities is included and correctly predicted. \\
\hline
3 & The prediction contains minor error(s) that do not affect the understanding of the relation. \\
\hline
2 & The prediction contains major error(s) that affect the understanding of the relation, while the relation can still be inferred to some extent. \\
\hline
1 & The prediction contains major error(s) that will mislead the understanding of the relation. \\
\bottomrule
\end{tabular}
\caption{Annotation guidelines excerpt.}
\vspace{-2mm}
\label{table:rating_scale}
\end{table}

\section{Implementation Details}
\label{app:implementation}

We employ the \texttt{fairseq} library\footnote{\url{https://github.com/pytorch/fairseq/tree/master/examples/bart}} to build the RelationBART model and adopt the key hyperparameters as suggested in \cite{lewis2020bart}. 
We manually set the learning rate as $5 \times 10^{-5}$ and batch-size of 1,024 tokens based on some preliminary experiments and the memory size of GPUs. 
We set the maximum reasoning length as 3 since the number of reasoning paths with hops $>3$ is very large and the quality of these paths is generally low.
For RelationBART-MP and reasoning path selection, we sample at most 5 reasoning paths with hops $\leq 3$. 
All the models were trained on NVIDIA Quadro RTX 5000 GPUs, and the training converged in 50 epochs. The training time of RelationBART-Vanilla, RelationBART-MP, and RelationBART-MP (Large) for one epoch with 3 GPUs are 80 minutes, 4 hours, and 7 hours respectively.

\section{Open Relation Modeling with Different Sizes of Training Data}
\label{app:small_data}

\begin{table}[h]
\begin{center}
\small
\setlength\tabcolsep{2pt} 
\begin{tabular}{l|c|c|c|c}
\toprule
\centering{\textit{100\%}} & \textbf{BL} & \textbf{R-L} & \textbf{MT} & \textbf{BS} \\
\midrule 
RelationBART-Vanilla (w/o PT) & 26.01 & 50.84 & 23.65 & 85.37  \\
RelationBART-Vanilla & 26.81 & 51.48 & 24.14 & 85.73 \\
\midrule[0.75pt]
\centering{\textit{10\%}} & \textbf{BL} & \textbf{R-L} & \textbf{MT} & \textbf{BS} \\
\midrule 
RelationBART-Vanilla (w/o PT) & 22.88 & 48.50 & 22.07 & 84.31 \\
RelationBART-Vanilla & 24.31 & 49.89 & 22.99 & 85.16  \\
\midrule[0.75pt]
\centering{\textit{1\%}} & \textbf{BL} & \textbf{R-L} & \textbf{MT} & \textbf{BS} \\
\midrule 
RelationBART-Vanilla (w/o PT) & 17.30 & 44.12 & 19.02 & 81.56 \\
RelationBART-Vanilla & 20.99 & 47.11 & 21.23 & 84.04 \\
\bottomrule
\end{tabular}
\caption{Results of open relation modeling with \textit{100\%}, \textit{10\%}, and \textit{1\%} training data.}
\label{table:relation_modeling_diff_train_size}
\vspace{-2mm}
\end{center}
\end{table}

From Table \ref{table:relation_modeling_diff_train_size}, we observe that when the training data become smaller, the performance of the version without pre-training decreases much faster than the one with pre-training.

\end{document}